\documentclass{mva_style}
\usepackage{anyfontsize}
\finalcopy 
\usepackage{graphicx}
\graphicspath{{pdf_figures/}}
\DeclareGraphicsExtensions{.pdf}
\usepackage{amsmath}
\usepackage[caption=false]{subfig}
\usepackage{url}
\usepackage{epsfig}
\usepackage{amssymb}
\usepackage[utf8]{inputenc}
\usepackage[english]{babel}
\usepackage [autostyle, english = american]{csquotes}
\usepackage{silence}
\usepackage[style=ieee, citestyle=numeric-comp, url=false, doi=false, isbn=false, eprint=false]{biblatex}
\AtEveryBibitem{\clearfield{pages}}
\AtEveryBibitem{\clearfield{volume}}
\AtEveryBibitem{\clearfield{number}}
\AtEveryBibitem{\clearfield{month}}

\MakeOuterQuote{"}

\usepackage{algorithm,tabularx}
\usepackage{algpseudocode}

\usepackage{arydshln}
\usepackage{multirow}
\usepackage{makecell}

\usepackage{xspace}

\usepackage{color}
\usepackage{tikz}

\usepackage[colorlinks=true,urlcolor=blue]{hyperref}

\algrenewcommand\algorithmicrequire{\textbf{Input:}}
\algrenewcommand\algorithmicensure{\textbf{Output:}}

\makeatletter
\newcommand{\multiline}[1]{%
	\begin{tabularx}{\dimexpr\linewidth-\ALG@thistlm}[t]{@{}X@{}}
		#1
	\end{tabularx}
}
\makeatother

\newcounter{algsubstate}

\makeatletter
\algnewcommand{\ProperState}[1]{\Statex \hskip\ALG@thistlm #1}
\makeatother

\DeclareMathOperator{\loss}{\mathcal{L}} 
\DeclareMathOperator{\gen}{\mathcal{G}}
\DeclareMathOperator{\dis}{\mathcal{D}}
\DeclareMathOperator{\enc}{\mathcal{E}}

\newcommand{\MSE}{\mathrm{MSE}}

\newcommand{\SSIM}{\mathrm{SSIM}}

\makeatletter
\DeclareRobustCommand\onedot{\futurelet\@let@token\@onedot}
\def\@onedot{\ifx\@let@token.\else.\null\fi\xspace}

\def\eg{\emph{e.g}\onedot}

\def\etc{\emph{etc}\onedot}

\makeatother

\newcolumntype{P}[1]{>{\centering\arraybackslash}p{#1}}

\newcommand{\superscript}[1]{\ensuremath{^{\textrm{#1}}}}

\begin{document}
\title{Predicting Next Local Appearance for Video Anomaly Detection}

\author{
  Pankaj Raj Roy\superscript{1} \qquad Guillaume-Alexandre Bilodeau\superscript{1} \qquad Lama Seoud\superscript{2}\\
  \superscript{1}LITIV \qquad \superscript{2}Institute of Biomedical Engineering\\
  Polytechnique Montr\'{e}al, Montr\'{e}al, Canada\\
  {\tt \{pankaj-raj.roy,gabilodeau,lama.seoud\}@polymtl.ca}\\
}

\maketitle

\section*{\centering Abstract}
\textit{
  We present a local anomaly detection method in videos. As opposed to most existing methods that are computationally expensive and are not very generalizable across different video scenes, we propose an adversarial framework that learns the temporal local appearance variations by predicting the appearance of a normally behaving object in the next frame of a scene by only relying on its current and past appearances. In the presence of an abnormally behaving object, the reconstruction error between the real and the predicted next appearance of that object indicates the likelihood of an anomaly. Our method is competitive with the existing state-of-the-art while being significantly faster for both training and inference and being better at generalizing to unseen video scenes.
}

\section{Introduction}
Video anomaly detection (VAD) is one of the behavioral recognition tasks that can help ensuring a safer society by detecting quickly various potential or ongoing incidents. Generally, VAD consists in identifying video frames containing spatio-temporal regions deviating from the normal behavior expected for a given scene. Since anomalies are rare and unpredictable, many researchers (\eg~\cite{Hasan2016,Luo2017a,Zhao2017,Ribeiro2018,Liu2018,Ionescu2019,Nguyen2019,Lu2020}) aim at identifying them as novelties in videos by relying only on normal known observations during training and by detecting anomalies as observations that fall outside of the known normal boundary.

Most of the recent VAD methods~\cite{Hasan2016,Luo2017a,Zhao2017,Ribeiro2018,Liu2018,Nguyen2019} are holistic. The normal pixel-level behaviors in videos are learned by training unsupervised generative neural networks minimizing the reconstruction and/or the prediction of the appearance/motion cues of the whole video frames. Color intensities and/or dense optical flows of video frames are widely used as appearance and/or motion cues respectively. In reconstruction-based methods~\cite{Hasan2016,Ribeiro2018}, these normal appearance and/or motion cues are learned through a convolutional auto-encoder (CAE) trained by minimizing the reconstruction error between the input and the predicted output. In the case of prediction-based approaches, which generally perform better compared to the reconstruction-based ones, CAEs~\cite{Zhao2017} or convolutional LSTMs~\cite{Luo2017a} with two decoders can reconstruct the input and predict the appearance of a different video frame by solely using some consecutive appearance cues, whereas U-Net frameworks~\cite{Liu2018,Nguyen2019} use both the appearance and motion cues. Even though they can perform VAD in real-time, most of these holistic methods are computationally heavy and tend to overfit on the background appearance in the training set. To alleviate the latter issue, model agnostic meta-learning (MAML~\cite{Finn2017}) was proposed in \cite{Lu2020} to extent a holistic VAD model beyond known video scenes. However, meta-learning scheme can be computationally very expensive when working with a high number of video scenes.

To overcome the issues of holistic methods, some researchers~\cite{Hinami2017,Ionescu2019} instead consider object-centric pixel-level features because they allow VAD systems to generalize better across different scenes. Indeed, each of these object-centric features corresponds to a spatio-temporal region occupied by one of the objects of interest (e.g. pedestrians, cyclists, cars, \etc), ignoring the non-object background information in videos that prevents VAD to generalize normal local behaviors to unseen videos. These approaches focus on the detection of object-centric local anomalies in videos, independently of other objects~\cite{Chen2015}, but actually they can also capture close-range interactions between objects allowing them to capture a broad variety of anomalies.

\begin{figure}[t]
	\centering
	\def\svgwidth{1\linewidth}
\begingroup%
  \makeatletter%
  \providecommand\color[2][]{%
    \errmessage{(Inkscape) Color is used for the text in Inkscape, but the package 'color.sty' is not loaded}%
    \renewcommand\color[2][]{}%
  }%
  \providecommand\transparent[1]{%
    \errmessage{(Inkscape) Transparency is used (non-zero) for the text in Inkscape, but the package 'transparent.sty' is not loaded}%
    \renewcommand\transparent[1]{}%
  }%
  \providecommand\rotatebox[2]{#2}%
  \newcommand*\fsize{\dimexpr\f@size pt\relax}%
  \newcommand*\lineheight[1]{\fontsize{\fsize}{#1\fsize}\selectfont}%
  \ifx\svgwidth\undefined%
    \setlength{\unitlength}{1531.50002307bp}%
    \ifx\svgscale\undefined%
      \relax%
    \else%
      \setlength{\unitlength}{\unitlength * \real{\svgscale}}%
    \fi%
  \else%
    \setlength{\unitlength}{\svgwidth}%
  \fi%
  \global\let\svgwidth\undefined%
  \global\let\svgscale\undefined%
  \makeatother%
  \begin{picture}(1,0.42335943)%
    \lineheight{1}%
    \setlength\tabcolsep{0pt}%
    \put(0,0){\includegraphics[width=\unitlength,page=1]{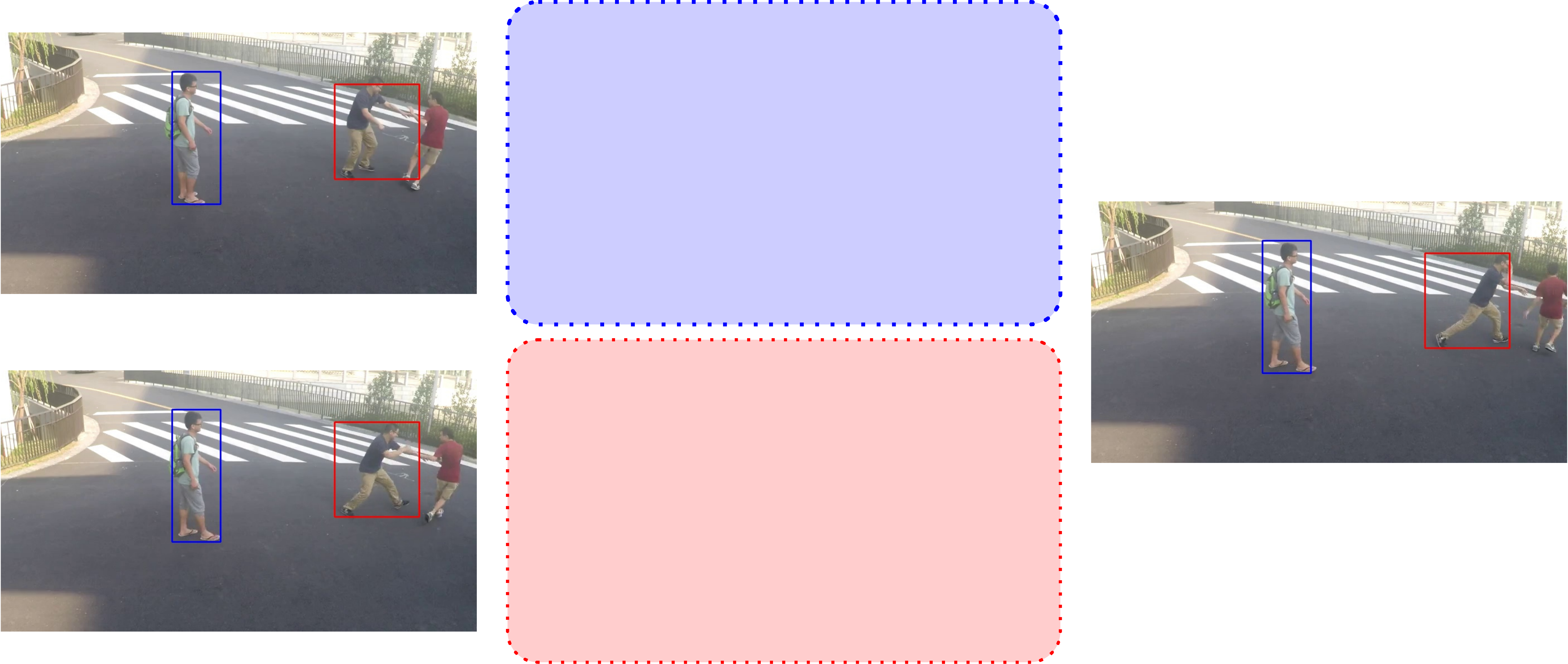}}%
    \put(0.07711878,0.24624256){\color[rgb]{1,1,1}\makebox(0,0)[lt]{\lineheight{1.25}\smash{\begin{tabular}[t]{l}$t-T$\end{tabular}}}}%
    \put(0.07645249,0.03112588){\color[rgb]{1,1,1}\makebox(0,0)[lt]{\lineheight{1.25}\smash{\begin{tabular}[t]{l}$t$\end{tabular}}}}%
    \put(0.77229762,0.13863276){\color[rgb]{1,1,1}\makebox(0,0)[lt]{\lineheight{1.25}\smash{\begin{tabular}[t]{l}$t+T$\end{tabular}}}}%
    \put(0,0){\includegraphics[width=\unitlength,page=2]{global_overview.pdf}}%
    \put(0.46827562,0.30814189){\color[rgb]{1,1,1}\makebox(0,0)[lt]{\lineheight{1.25}\smash{\begin{tabular}[t]{l}$\gen$\end{tabular}}}}%
    \put(0,0){\includegraphics[width=\unitlength,page=3]{global_overview.pdf}}%
    \put(0.46827565,0.08972857){\color[rgb]{1,1,1}\makebox(0,0)[lt]{\lineheight{1.25}\smash{\begin{tabular}[t]{l}$\gen$\end{tabular}}}}%
    \put(0,0){\includegraphics[width=\unitlength,page=4]{global_overview.pdf}}%
  \end{picture}%
\endgroup%

	\caption{Overview of our method. For each object, the generator $ \gen $ predicts the appearance of that object in the next frame $ t + T $ by using its appearances in the past $ t - T $ and current $ t $ frames. For an anomaly (person fighting), $ \gen $ is expected to produce bad reconstruction (as shown in red).}
	\label{fig:our_method_overview}
\end{figure}

Because of their advantages over holistic models, we propose to use  object-centric appearances extracted with a pretrained object detector for training a VAD framework that can detect local anomaly. We also want a light-weight VAD solution that can quickly and effectively detect local anomalies with just a few video frames. As opposed to some existing methods~\cite{Liu2018,Ionescu2019,Nguyen2019}, we do not use any motion features (\eg optical flows) mainly due to the fact that they are noisy for low resolution/poor quality videos. Moreover, the explicit use of motion adds another layer of complexity that can make the VAD framework less generalizable to other videos with different viewpoints.

Unlike the framework proposed in \cite{Ionescu2019} where each CAE model learns normal features independently from each other, we propose a next local appearance prediction network (NLAPnet) comprising of a single generator that follows a U-Net architecture with skip connections to learn to predict the next local appearance of a normally behaving object by using only the past and the current local appearances of the corresponding object in the video frames. Thus, contrarily to the CAE models in \cite{Ionescu2019}, our generator explicitly learns the short temporal appearance variations of a given object which enables better characterization of normal behavior. Inspired by \cite{Liu2018,Nguyen2019} in which the generator is trained as a generative adversarial network (GAN~\cite{Goodfellow2020}), an adversarial loss is also added during training in which a discriminator learns jointly with the generator to separate the real against the generated images. During inference for a given object, we rely on the structural similarity index measure (SSIM~\cite{Wang2004}) between the real and the predicted image for producing an anomaly score.

Our contributions are: 1) Unlike previous methods, ours explicitly learns the temporal appearance variations by predicting the next local appearance using just a few frames and we show that this approach gives competitive results, 2) our method can perform real-time VAD when using a pretrained DLA backbone for object detection, and 3) because it does not rely on optical flow and it is object-centric, our method shows very good generalization capabilities on new scene as demonstrated by a few-shot scene adaptation study.

\section{Proposed Method}
Assuming that local video anomalies are caused by objects in the scene, we propose to adversarially train, with a discriminator, a generative model that learns to predict the appearance of objects behaving normally in the next frame given its appearances in the past and current frames. To reduce the computation cost, we define past ($ a_i^{t - T} $), current ($ a_i^{t} $) and next ($ a_i^{t + T} $) object-centric appearances as the gray-scaled pixel-level intensity images of an object ($ i $) in the past ($ t - T $), current ($ t $) and next ($ t + T $) frames respectively. Hence, each object in a given scene at a given time has an associated appearance triplet ($ <a_i^{t - T}, a_i^{t}, a_i^{t + T}> $) which slides temporally through the video frames. The frame gap $ T = 3 $ is empirically selected to ensure significant local changes in appearance while having a small spatial displacement to avoid the need to track the objects. After learning with normal behaviors, the neural network will produce bad reconstructions for the next appearance of objects with abnormal behavior, thus indicating the likelihood of anomalies.

\begin{figure}[t]
	\centering
	\def\svgwidth{1\linewidth}
	\begin{footnotesize}
		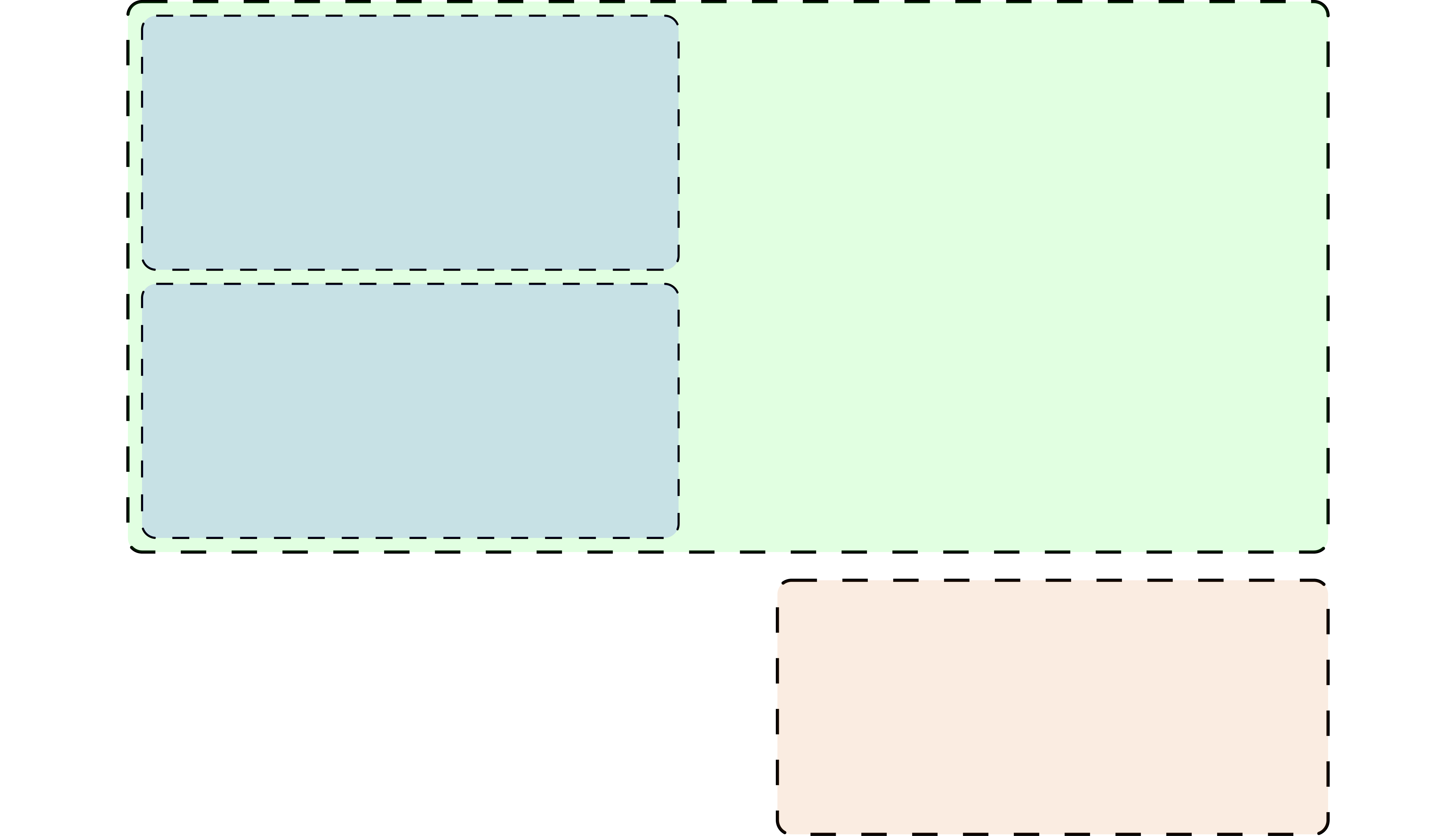
	\end{footnotesize}
	\caption{Overview of our proposed next local appearance prediction network (NLAPnet). The encoders $ \enc_{p} $ and $ \enc_{c} $ of the generator $ \gen $ take the past and current appearances $ a_i^{t - T} $ and $ a_i^{t} $ respectively. The decoder learns to predict the next appearance $ a_i^{t + T} $. The discriminator $ \dis $ learns to discriminate the real and the predicted images.}
	\label{fig:our_network_overview}
\end{figure}

\subsection{Object-Centric Appearance Extraction}
Since our method requires object-centric appearances in videos, we need a pretrained multi-class object detector (MOD) to estimate the bounding box locations of all the objects of interest for each of the video frames. For this, we used the CenterNet~\cite{Zhou2019c} detector, for his state-of-the-art (SOTA) performance, with pretrained weights on MS-COCO~\cite{Lin2014} for detecting objects within 81 different classes. To extract the appearance images of a given object, we simply crop the past ($ t - T $), the current ($ t $) and the next ($ t + T $) video frames using the bounding box coordinates of that object in the current ($ t $) frame.

\subsection{Next Local Appearance Prediction Network}
NLAPnet predicts the next local appearance of an object using a generator $ \gen $ that learns to predict $ a_i^{t + T} $ from a pair of object crops $ <a_i^{t - T}, a_i^{t}> $. In order to solve the issue of vanishing gradients while capturing better fine-grained details, our generator $ \gen $ is formed by two encoders that are linked to one decoder with skip-connections~\cite{Ronneberger2015}.  The encoders share the same weights and are composed of multiple 2-strided convolutional layers, while the decoder is composed of multiple 2-strided transpose convolutional layers. A discriminator $ \dis $ is also added to impose a generative adversarial constraint by learning to differentiate the real $ a_i^{t + T} $ against the generated one $ \hat{a}_i^{t + T} $ by $ \gen $. To increase generative performance, our model follows the adversarial framework in \cite{Liu2018} by using PatchGAN~\cite{Isola2017} with the Least Square GAN~\cite{Mao2017} which makes $ \dis $ a discriminative feature generator that is composed of multiple 2-strided convolutional layers and learns to output a feature matrix of confidence values ranging from one (for real images) to zero (for generated images).

For a given pair of $ <a_i^{t - T}, a_i^{t}> $ of an object $ i $ in a video frame $ t $, the quality of the prediction by $ \gen $ is evaluated through the generative reconstruction loss $ \loss_{\gen} $, the generative adversarial loss $ \loss_{adv}^{\gen} $ and the discriminative adversarial loss $ \loss_{adv}^{\dis} $, which are given by:
\begin{align}
	\loss_{\gen} &= \frac{1}{2} \left( 1 - \SSIM \left( a_i^{t + T}, \gen \left( a_i^{t - T}, a_i^{t} \right) \right) \right) \\ \label{equ:vad}
	\loss_{adv}^{\gen} &= \sum_{x,y} \left( 1 - \dis \left( \gen \left( a_i^{t - T}, a_i^{t} \right) \right)_{x,y} \right)^2 \\ 
	\begin{split}
		\loss_{adv}^{\dis} &= \frac{1}{2} \left[ \sum_{x,y} \left( 1 - \dis \left( a_i^{t + T} \right)_{x,y} \right)^2 \right. \\
		&\qquad \left. + \sum_{x,y} \left( \dis \left( \gen \left( a_i^{t - T}, a_i^{t} \right) \right)_{x,y} \right)^2 \right] \text{,}
	\end{split}
\end{align}
where $ x $ and $ y $ denotes the patch coordinates of the discriminative feature for computing the mean squared errors ($ \MSE $). During training, $ \dis $ and $ \gen $ update their weights alternately : $ \dis $ first minimizes $ \loss_{adv}^{\dis} $ and then $ \gen $ minimizes $ \loss_{\gen} + \loss_{adv}^{\gen} $. For $\loss_{\gen}$, we found that the structural similarity index measure (SSIM~\cite{Wang2004}) between gray-scaled intensity images yields better overall VAD performance.

\subsection{Video Anomaly Detection}
During inference, we simply use $ \loss_{\gen} $ which employs $ \SSIM $ as described in eq.~\ref{equ:vad} for producing a region-level anomaly score $ s_i^t $ for a triplet $ <a_i^{t - T}, a_i^{t}, a_i^{t + T}> $ of an object $ i $ at a time $ t $. Anomaly detection decisions are taken at frame-level with provisions to reduce noise. To get the frame-level anomaly score $ s^t $, we take the highest corresponding region-level anomaly score. To mitigate the issue of missing detection due to occlusion, we temporally smooth the frame-level anomaly scores with a Gaussian filter.

\section{Experiments}

\subsection{Experimental setup}
To validate our proposed method, we used three publicly available VAD benchmarks with training videos assumed as being normal: UCSD Pedestrian~\cite{Mahadevan2010}, CUHK Avenue~\cite{Lu2013} and ShanghaiTech~\cite{Luo2017}. The UCSD Pedestrian benchmark is divided into two datasets (Ped1 and Ped2), where both share the same abnormalities in videos consisting of vehicles, cyclist, skateboarders and wheelchairs going on pedestrian pathways, the CUHK Avenue dataset contains anomalies that are based on some irregular human activities like running, loitering and throwing/leaving carried objects, and ShanghaiTech (ST) is composed of multiple videos having different scenes with varying types of abnormal activities like cycling, fighting, robbing, \etc.




To measure performance, we employ the frame-level area under the ROC curve, following the evaluation protocol in \cite{Liu2018}. To study the effect of MOD on local VAD, we consider two different CenterNet~\cite{Zhou2019c} backbones, namely 1) deep layer aggregation with 34 layers (DLA~\cite{Yu2018}) and 2) hourglass with 104 layers (HG~\cite{Law2020}). The former backbone is less accurate than the latter one, but enables fast real-time detection. When conducting experiments related to scene adaptation and for the ablation study, we used the HG backbone.

\begin{table}[t]
	\centering
	\caption{Frame-level AUC results (in \%) of our method on four datasets. * Means that the results were obtained by our implementation of the method. The best results are shown in boldface.}
	\label{tab:results}
	\begin{tabular}{l|cccc}
		Method                  		&   Ped1   &   Ped2   &   Ave    &   ST     \\ \hline
		CAE~\cite{Hasan2016}     		& $ 81.0 $ & $ 90.0 $ & $ 70.2 $ & $ 60.9 $ \\
		ConvLSTM-AE~\cite{Luo2017a} 	& $ 75.5 $ & $ 88.1 $ & $ 77.0 $ &     -    \\
		STAE-OF~\cite{Zhao2017}     	& $ \mathbf{87.1} $ & $ 88.6 $ & $ 80.9 $ &     -    \\
		Deep CAEs~\cite{Ribeiro2018}   	& $ 56.9 $ & $ 84.7 $ & $ 77.2 $ &     -    \\
		FFP~\cite{Liu2018}       		& $ 83.1 $ & $ 95.4 $ & $ 84.9 $ & $ 72.8 $ \\
		MAC~\cite{Nguyen2019}   		&     -    & $ 96.2 $ & $ \mathbf{86.9} $ &     -    \\
		Few-shot~\cite{Lu2020}   		& $ 86.3 $ & $ 96.2 $ & $ 85.8 $ & $ 77.9 $ \\ \hline
		OC-CAEs-dla*~\cite{Ionescu2019} & $ 77.4 $ & $ 95.5 $ & $ 80.3 $ & $ 79.3 $ \\
		OC-CAEs-hg*~\cite{Ionescu2019}  & $ 79.6 $ & $ 96.4 $ & $ 82.2 $ & $ 80.5 $ \\ \hline
		NLAPnet-dla              		& $ 81.1 $ & $ 96.3 $ & $ 81.3 $ & $ 82.0 $ \\
		NLAPnet-hg               		& $ 82.3 $ & $ \mathbf{97.2} $ & $ 85.4 $ & $ \mathbf{82.5} $ \\ \hline
	\end{tabular}
\end{table}
\begin{table}[t]
	\centering
	\caption{Comparison between frame-level AUC results (in \%) on three target datasets using pretrained weights on ST. Met.: Method, Tr.: Training scheme. PO: use pretrained weights only, FT: fine-tune the pretrained weights on the target dataset and MT: meta-training on ST. *: results were obtained by our implementation of the method. $\dagger$: shots are not used with PO. Best results are shown in boldface.}
	\label{tab:adapt_results}
	\begin{tabular}{l|l|l|ccc}
		Target                  		&   Met.& Tr.   	   	&  1-shot  &  5-shot  &  10-shot \\ \hline
		\multirow{5}{*}{Ped1}     		& ~\cite{Lu2020} & PO 	&  $73.1\dagger$ & \multicolumn{2}{c}{}  \\
		& ~\cite{Lu2020} 		& FT	& $ 77.0 $ & $ 77.9 $ & $ 78.2 $ \\
		& ~\cite{Lu2020} 		& MT	& $ \mathbf{80.6} $ & $ \mathbf{81.4} $ & $ \mathbf{82.4} $ \\ \cdashline{2-6}
		& *~\cite{Ionescu2019} 	& PO  	&   $71.6\dagger$ & \multicolumn{2}{c}{} \\
		& *~\cite{Ionescu2019} 	& FT  	& $ 72.1 $ & $ 74.3 $ & $ 77.2 $ \\ \cdashline{2-6}
		& (ours) 				& PO	&   $74.4\dagger$ & \multicolumn{2}{c}{} \\
		& (ours) 				& FT	& $ 76.3 $ & $ 77.1 $ & $ 79.3 $ \\ \hline
		\multirow{5}{*}{Ped2}     		& ~\cite{Lu2020} & PO 			&   $82.0\dagger$ & \multicolumn{2}{c}{} \\
		& ~\cite{Lu2020} 		& FT 	& $ 85.6 $ & $ 89.7 $ & $ 91.1 $ \\
		& ~\cite{Lu2020} 		& MT	& $ 91.2 $ & $ 91.8 $ & $ 92.8 $ \\ \cdashline{2-6}
		& *~\cite{Ionescu2019} 	& PO  	&   $90.3\dagger$ & \multicolumn{2}{c}{} \\
		& *~\cite{Ionescu2019} 	& FT  	& $ 92.2 $ & $ 93.1 $ & $ 94.2 $ \\ \cdashline{2-6}
		& (ours) 				& PO	  & $95.9\dagger$ & \multicolumn{2}{c}{} \\
		& (ours)				& FT	& $ \mathbf{96.5} $ & $ \mathbf{96.8} $ & $ \mathbf{97.5} $ \\ \hline
		\multirow{5}{*}{Ave}     		& ~\cite{Lu2020} & PO 			&   $71.4\dagger$ & \multicolumn{2}{c}{} \\
		& ~\cite{Lu2020} 		& FT	& $ 75.4 $ & $ 76.5 $ & $ 77.8 $ \\
		& ~\cite{Lu2020} 		& MT	& $ 76.6 $ & $ 77.1 $ & $ 78.8 $ \\ \cdashline{2-6}
		& *~\cite{Ionescu2019} 	& PO  	&  $74.7\dagger$ & \multicolumn{2}{c}{} \\
		& *~\cite{Ionescu2019}  & FT	& $ 75.7 $ & $ 76.5 $ & $ 77.1 $ \\ \cdashline{2-6}
		& (ours) 				& PO	&   $78.8\dagger$ & \multicolumn{2}{c}{} \\
		& (ours)				& FT	& $ \mathbf{79.5} $ & $ \mathbf{81.1} $ & $ \mathbf{82.3} $ \\ \hline
	\end{tabular}
\end{table}
\begin{table}[t]
	\centering
	\caption{Ablation study AUC results (in \%) on ST. $ \enc_{p} $ and $ \enc_{c} $: encoders of the generator $ \gen $ for past and current local images respectively. SC: with skip-connections. $ adv $: $ \gen $ is trained adversarially with a discriminator $ \dis $. Best results are shown in boldface.}
	\label{tab:ablation_study}
	\begin{tabular}{c|cccccc}
		$ \enc_{p} $ 	& \checkmark &     ---    & \checkmark & \checkmark & \checkmark & \checkmark \\
		$ \enc_{c} $ 	&     ---    & \checkmark & \checkmark & \checkmark & \checkmark & \checkmark \\
		SC  			&     ---    &     ---    &     ---    & \checkmark &     ---    & \checkmark \\
		$ adv $  		&     ---    &     ---    &     ---    &     ---    & \checkmark & \checkmark \\ \hline
		AUC            	&  $ 75.3 $  &  $ 75.5 $  &  $ 76.7 $  &  $ 80.8 $  &  $ 76.2 $  &  $ \mathbf{82.2} $
	\end{tabular}
\end{table}
\begin{table}[t]
	\centering
	\caption{Training (Tr) time in hours and inference (Inf) speed in frame per seconds of our method and others on ST.}
	\label{tab:running_times}
	\begin{tabular}{l|c|cc|cc}
				& \cite{Liu2018} & dla~\cite{Ionescu2019} & hg~\cite{Ionescu2019} & dla (ours) & hg (ours) \\ \hline
		Tr		& $ 48 $ & $ 10 $ & $ 10 $ & $ \mathbf{3} $ & $ \mathbf{3} $ \\
		Inf 	& $ 25 $ & $ 25 $ & $ 11 $ & $ \mathbf{34} $ & $ 13 $
	\end{tabular}
\end{table}

\subsection{Results and Discussion}
\textbf{Comparison with SOTA methods:} We first compared our proposed method using two different MOD backbones (NLAPnet-dla and NLAPnet-hg) with several existing SOTA methods~\cite{Hasan2016,Luo2017a,Zhao2017,Ribeiro2018,Liu2018,Nguyen2019,Ionescu2019,Lu2020}, as shown in table~\ref{tab:results}. Note that we present only the results from our version of the implementation of \cite{Ionescu2019} using the same two MOD backbones (OC-CAEs-dla and OC-CAEs-hg) so that all results are obtained following the same evaluation protocol for a rigorous comparison. Overall, results show that our proposed method, using either one of the two MOD backbones, significantly outperforms other existing approaches on the most challenging dataset, ST, as well as on Ped2, while being competitive on other datasets. However, our method does not perform as well on Ped1 which has a significantly lower resolution, which may affect the learning of normal behavior for predicting frames. Nevertheless, we can see that a better performing MOD backbone helps increasing the performance on these datasets, especially on Avenue, with heavily occluded areas. Still, our results with NLAPnet-dla are competitive and are obtained at real-time speed. With NLAPnet-hg, our performances are similar or better than holistic methods~\cite{Hasan2016,Luo2017a,Zhao2017,Ribeiro2018,Liu2018,Nguyen2019,Lu2020}, except for Ped1. This shows the benefit of an object-centric approach in VAD. 

\textbf{Performance in Scene Adaptation: }
Since the non-object background information is not used in our proposed framework, we are more capable of adapting NLAPnet to a different unseen video scene with similar anomalies. To demonstrate that, we employed a similar protocol as \cite{Lu2020} by fine-tuning our pretrained model with only $ K $ randomly selected frames per video ($ K $-shot) in the training set of the target datasets: Ped1, Ped2 and Avenue (Ave). As in \cite{Lu2020}, we use ST as the source dataset for pretraining our model since it has various scenes with similar anomalies as in other datasets. Moreover, to further compare the effectiveness of our proposed framework for scene adaptation, we include results with the existing object-centric method~\cite{Ionescu2019} by fine-tuning the CAE models on the target dataset. Results are shown in table~\ref{tab:adapt_results}.

Results show that our method outperforms the model agnostic meta-learning (MAML~\cite{Finn2017}) framework that was used as an holistic method for the Ped2 and Avenue datasets. We also note a significant increase in performance when using our framework compared to the object-centric method of \cite{Ionescu2019}. Interestingly, we can see that using pretrained weights on ST can even improve the performance on Ped2 compared to the one that is trained only on Ped2 (see table~\ref{tab:results}). However, for Ped1, since the appearances of crowded objects are of low pixel resolution, similarly to~\cite{Ionescu2019}, our method does not perform as well compared to the holistic model~\cite{Lu2020}.

\textbf{Ablation Study:} 
In this experiment, we validate the crucial parts of NLAPnet using the ST dataset. We test the effects of the two encoders (for past and current images), the skip-connections and the adversarial loss $\loss_{adv}^{\dis}$. As we can see in table~\ref{tab:ablation_study}, results illustrate the importance of using both encoders with skip-connections when training the model in an adversarial manner.

\textbf{Running Time:}
We used Python 3 and Keras 2 with TensorFlow binding to implement our proposed method\footnote{Code: \url{https://github.com/proy3/NLAP-net_VAD}.} on a Intel i7-8700 machine with 16 GB RAM using Nvidia RTX 2080 GPU. Table~\ref{tab:running_times} gives the approximated training time and inference speed of NLAPnet, of  \cite{Liu2018} and of our implementation of \cite{Ionescu2019} using DLA or HG MOD backbone. We can see that our method is significantly faster than \cite{Ionescu2019} while having competitive VAD performances.

\section{Conclusion}
To conclude, this paper proposes an adversarial framework that learns to predict the local appearance of a normally behaving object in the next video frame by using only the appearances of that object in the past and current frames. Results on four public benchmarks demonstrate the effectiveness of our method with competitive performance in VAD while at the same being faster at inference, light-weight and capable to better adapt to unseen video scenes than other methods.

\newpage
\section*{Acknowledgments}
This research was supported by grants from IVADO and NSERC funding programs.
\printbibliography

\end{document}